\documentclass[conference]{IEEEtran}
\IEEEoverridecommandlockouts
\usepackage{cite}
\usepackage{amsmath,amssymb,amsfonts}
\usepackage{graphicx}
\usepackage{textcomp}
\usepackage{xcolor}
\usepackage{nameref}
\usepackage{algorithm}
\usepackage{algpseudocode}
\usepackage{makecell}

\def\BibTeX{{\rm B\kern-.05em{\sc i\kern-.025em b}\kern-.08em
    T\kern-.1667em\lower.7ex\hbox{E}\kern-.125emX}}

\makeatletter
\newcommand{\linebreakand}{%
  \end{@IEEEauthorhalign}
  \hfill\mbox{}\par
  \mbox{}\hfill\begin{@IEEEauthorhalign}
}
\makeatother

\begin{document}

\title{\vspace{0.5cm} TeslaCharge: Smart Robotic Charger Driven by Impedance Control and Human Haptic Patterns
}

\author{\IEEEauthorblockN{Oussama Alyounes}
\IEEEauthorblockA{\textit{Skolkovo Institute of Science  
and Technology}\\
Moscow, Russia \\
oussama.alyounes@skoltech.ru}
\and
\IEEEauthorblockN{Miguel Altamirano Cabrera}
\IEEEauthorblockA{\textit{Skolkovo Institute of Science 
and Technology}\\
Moscow, Russia \\
miguel.altamirano@skoltech.ru} 

\linebreakand
\IEEEauthorblockN{Dzmitry Tsetserukou}
\IEEEauthorblockA{\textit{Skolkovo Institute of Science
and Technology}\\
Moscow, Russia \\
d.tsetserukou@skoltech.ru}

}

\maketitle

\begin{abstract}

The growing demand for electric vehicles requires the development of automated car charging methods. At the moment, the process of charging an electric car is completely manual, and that requires physical effort to accomplish the task, which is not suitable for people with disabilities. Typically, the effort in the research is focused on detecting the position and orientation of the socket, which resulted in a relatively high accuracy, $\pm 5 \: mm $ and $\pm 10^o$. However, this accuracy is not enough to complete the charging process. In this work, we focus on designing a novel methodology for robust robotic plug-in and plug-out based on human haptics, to overcome the error in the position and orientation of the socket. Participants were invited to perform the charging task, and their cognitive capabilities were recognized by measuring the applied forces along with the movement of the charger. Three controllers were designed based on impedance control to mimic the human patterns of charging an electric car. The recorded data from humans were used to calibrate the parameters of the impedance controllers: inertia $M_d$, damping $D_d$, and stiffness $K_d$. A robotic validation was performed, where the designed controllers were applied to the robot UR10. Using the proposed controllers and the human kinesthetic data, it was possible to successfully automate the operation of charging an electric car.

\end{abstract}


\section{Introduction}

\begin{figure}[t!]
    \centering
    \includegraphics[width=0.47\textwidth]{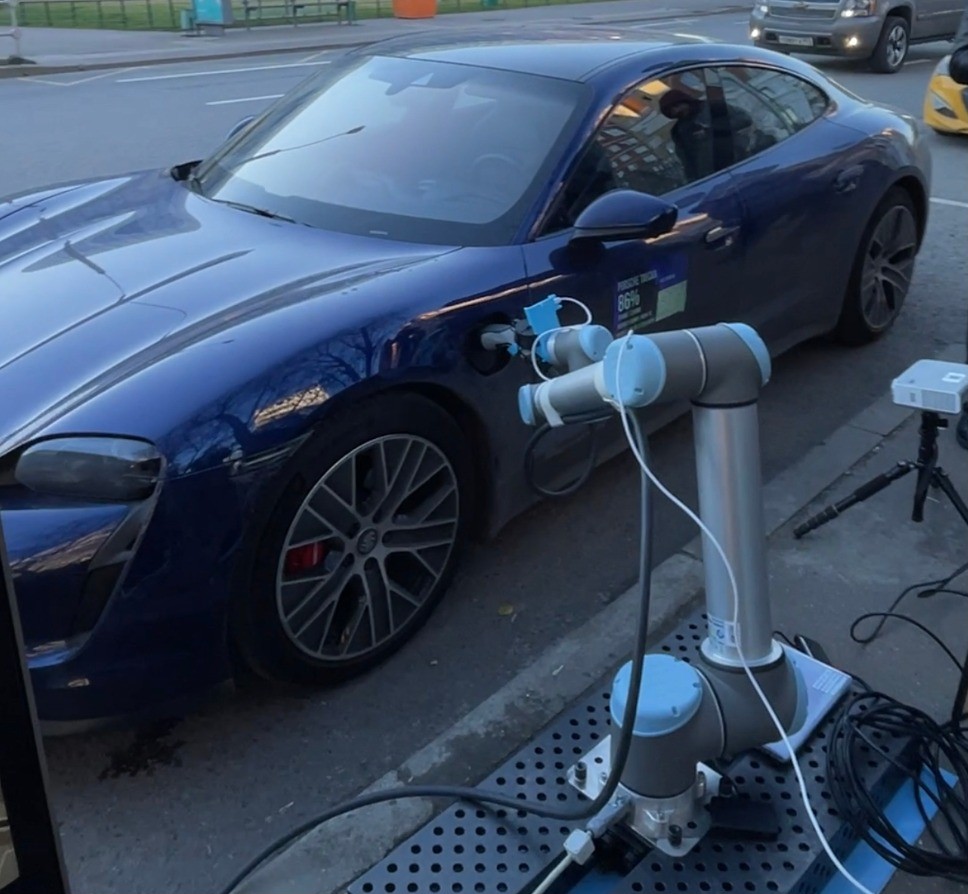}
    \vspace{-0,25 cm}
    \caption{Robotic manipulator plugging the charger inside the car socket.}
    
    \label{fig:car&charger}
    \vspace{-0.5cm}
\end{figure}

Electric vehicles (EVs) are widely seen as the future of zero-emission transportation, representing a significant step forward in the development of eco-friendly mobility \cite{AutomotiveWorld}. Over one million EVs were sold in 2017, and more than 10 million in 2022 \cite{IEA}, with an expectation in the sales increase reaching 30\% by 2030 \cite{electricCarsSales}. Furthermore, there is an anticipation that by the year 2035, every newly registered car in Europe will be of zero-emission type \cite{zero_emission_2035}. As EVs undergo development with a focus on improving performance, the process of charging them remains entirely manual, where people are responsible for connecting the charging cable to the socket \cite{intro_technologies}. A survey of people with impairments was conducted by the British Research Institute for Disabled Consumers (RiDC) showing that only 25\% had the desire to use EVs \cite{RiDC}. This is due to the inadequacy of charging stations, which are often situated in inconvenient locations, requiring physical effort to access and connect the charger. The survey also showed that this number increases up to 61\% when the charging infrastructures were promised to be enhanced. The growing demand for EVs, with the goal of satisfying consumer requirements, require the creation of innovative and user-friendly charging methods.

The process of EVs charging can be likened to the peg-in-hole assembly task, a concept that has been extensively researched and employed in various applications, including the assembly of large-scale components \cite{Wan_large_scale} and screw plug-in \cite{screws_insertion}. Robotic assembly tasks demand a significant level of repeatability and adaptability. These qualities are often attained through the precise programming of specific positions and trajectories \cite{HU2011715}. Force-torque (FT) sensors were implemented in the process to increase the success rate of inserting \cite{pih_example1}. 

Human-inspired compliant was also designed with an FT sensor to detect the direction of the movement of the peg \cite{pih_example4}. However, human behavior was only studied by observation, with no attempt to record and process the data to achieve an optimal controller.

This paper presents a methodology for smart robotic chargers during plug-in and plug-out phases based on human haptic patterns. The experiment on human data acquisition is presented in Section \ref{sec:Data_Acquisition_Experiment}, where the human patterns were recognized using an FT sensor and the movement of the charger. In Section \ref{sec:Controller_Design}, a controller for the plug-in and out based on impedance controller is presented, where human data is considered to determine the coefficients of the rotational and linear controllers. Section \ref{sec:robotic_evaluation} shows the use of the human study to determine the parameters of the impedance controllers implemented to perform the operation by a robotic arm from Universal Robots. The experiments with the robot are further described in Section \ref{sec:robotic_evaluation}.

\section{Related Work}

When humans engage in assembly tasks, they primarily rely on two key senses. The visual sense to estimate the locations of the parts and make approximate matches, while the kinesthetic sense is used to complete the assembly by fitting the peg into the hole \cite{Abdullah2015}. Itabashi and his team developed a method to replicate human peg-in-hole skills by analyzing impedance parameters derived from human demonstrations \cite{Itabashi}. 
By analyzing human motion conditions, it is possible to build manipulator control architectures based on human skills for different applications \cite{Zhu18, screwingSk}.

To perform the EV charging automation task, we need a controller that achieves a dynamic interaction between a manipulator (charger) and its environment (socket). Impedance and admittance controllers are two types of controllers that regulate such sort of interactions  \cite{impedance-summary}.  
They regulate the interaction between the force and the motion (position or speed) \cite{admittance-vs-impedance}. Our task requires us to change the motion of the charger depending on the forces measured from the FT sensor, which can be achieved using impedance control. Applying an impedance controller requires adjusting its parameters: inertia, damping, and stiffness. Several studies with different approaches were introduced to adjust these parameters, such as depending on reinforcement learning (RL) \cite{impedance-learning-variables-RL}, and learning from demonstration \cite{impedance-learning-variables-demonstration}.

The full automation of the process of EV charging incorporates three phases: the localization of the electric port, the plug-in of the electric charger in the socket, and the plug-out. The initial phase is a computer vision (CV) task focused on detecting the socket's position and orientation with minimal errors \cite{Zou2019}. The primary challenges of the plug-in process are the small clearances between the socket and peg, along with aligning numerous pins with differing radii. M. Jokesch et al. analyzed the error in the localization of the socket \cite{jokesch2016generic} and an impedance controller was applied to compensate the error in 4 out of 6 degrees of freedom (DOF) depending on the measured forces at the tool center point (TCP). However, this study used the robot KUKA LWR iiwa 7 R800 with seven DOF that allowed to insert the charger depending on the compliance of the KUKA robot. X. Lv et al. proposed an automatic system guided by a camera and an FT sensor to support the successful completion of the plug-in process while charging an electric car, reporting a 92\% success rate in the operation \cite{surf_template_match_FT}. However, the orientation of the socket was precisely known during the plug-in, and it followed a unique strategy to plug in the connector.

\begin{figure}
 \centering
 \includegraphics[width=0.36\textwidth]{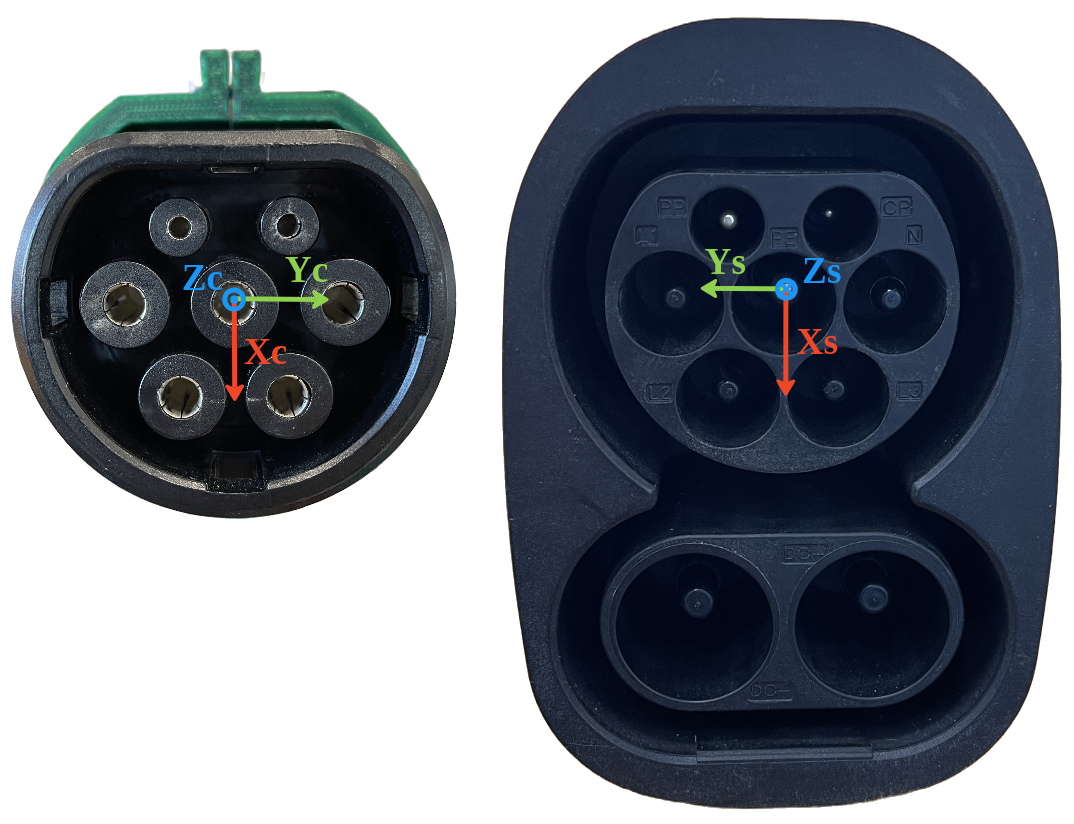}
 \caption{Socket and charger of type 2 used in the work.}
 \label{fig:socket&charger}
 \vspace{-0.35cm}
\end{figure}

\section{System Overview}

The general goal of this work is to automate the process of charging an electric car. This paper presents a novel methodology for robust robotic plug-in and plug-out development during the charging process using human data. It is assumed that the first phase of localizing the socket is already done, and the focus of this work is on compensating the position and orientation errors. We integrated an FT sensor to support the process of inserting the charger. To achieve a successful plug-in and plug-out, we measure the forces and torques applied to the charger to correct the position and orientation of the charger during the operation using impedance control. The parameters of the impedance controller were obtained by analyzing human patterns. An electric car charger and socket of type 2 were used in this study as shown in Fig. \ref{fig:socket&charger}, where it can be seen that this model is composed of seven different pins. Changing to another type of the socket/charger will affect the controller parameters but new parameters can be obtained by applying the same methodology presented in this paper.

The electric car charger was attached to a 6-DOF FT sensor 150-S from Robotiq, which works in the measuring ranges $\pm 150$  $N$ and $\pm 15$ $Nm$ for the forces and torques, respectively, and has an output rate of $100$  $Hz$ \cite{robotiq_sensor}. The FT sensor was used in the \nameref{sec:Data_Acquisition_Experiment} (Section \ref{sec:Data_Acquisition_Experiment}) to measure the forces and torques that the participants apply during the plug-in and out of the charger, and in the \nameref{sec:robotic_evaluation} (Section \ref{sec:robotic_evaluation}) to measure the interacted forces and torques between the charger and the socket.

For the \nameref{sec:Data_Acquisition_Experiment} (Section \ref{sec:Data_Acquisition_Experiment}), a charger holder was designed, 3D printed, and assembled with the FT sensor and the charger. The holder enables the user to carry the charger while plugging it in and out of the socket. Two ArUco markers were attached to the charger setup with the goal of tracking and analyzing the position of the charger while the participants were carrying out the experiment. Furthermore, a camera of type Logitech HD 1080p C930e was mounted on a stand in a fixed position for the whole experiment. The camera detects the positions of the two ArUco markers in order to process the charger's movements. Two ArUco markers were used to cover failure cases when one marker was not detected.

\begin{figure}[h]
 \centering
 \includegraphics[width=0.45\textwidth]{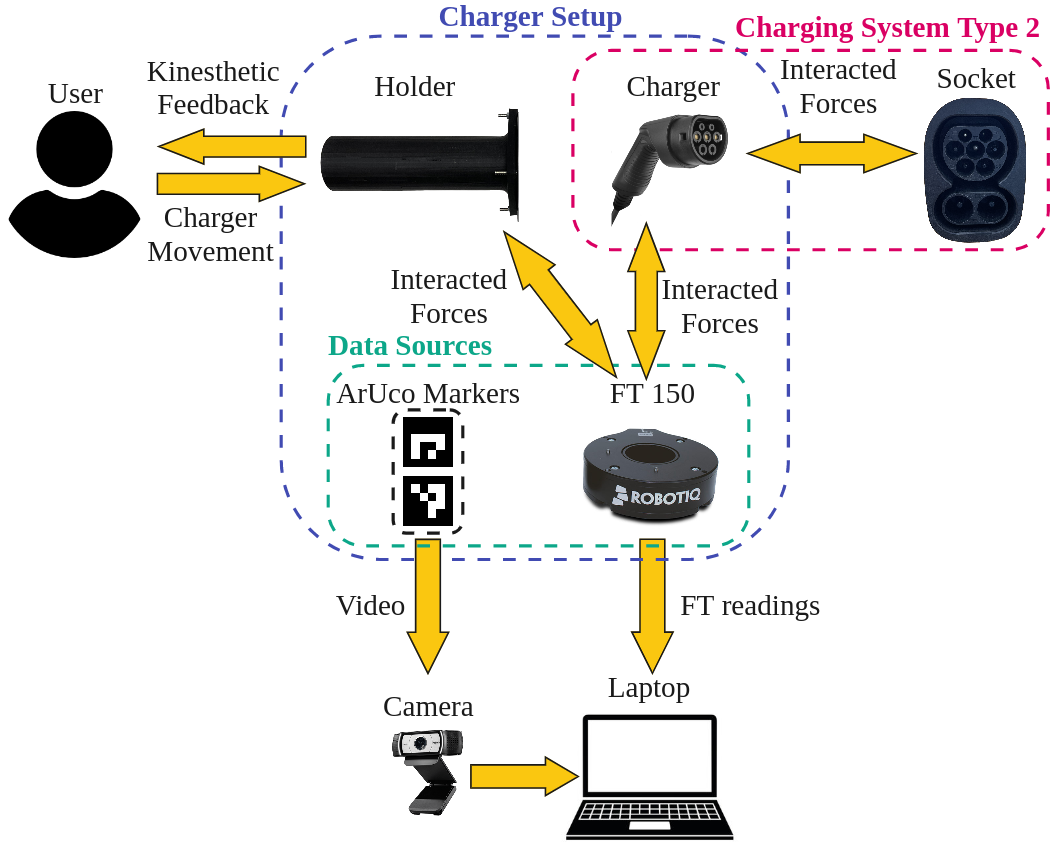}
 \caption{System architecture used for the human study experiment.}
 \label{fig:system_architecture_user}
\end{figure}

The system architecture for the \nameref{sec:Data_Acquisition_Experiment} is depicted in Fig. \ref{fig:system_architecture_user}. This figure shows the interacted forces that are being transferred from the socket towards the user's hands and the user is moving the charger to complete the task while readings from the FT sensor the position of the ArUco markers are being recorded for further analysis.

\begin{figure}[h]
 \centering
 \includegraphics[width=0.45\textwidth]{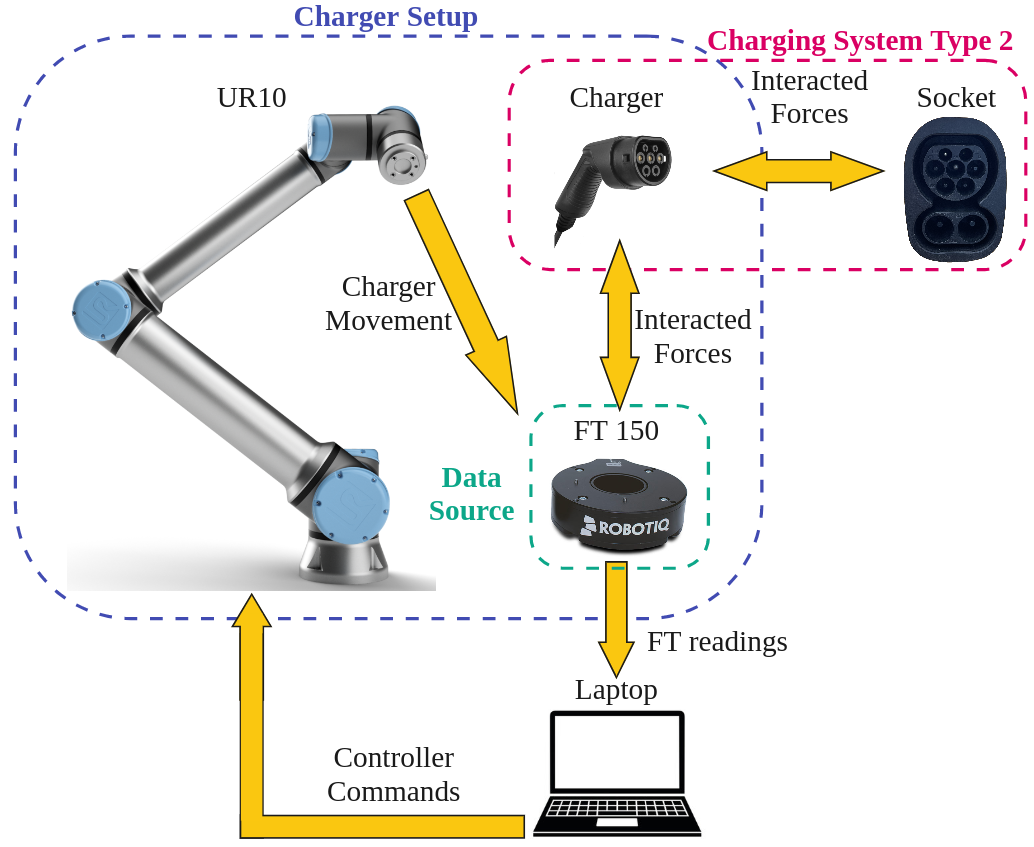}
 \caption{System architecture used in the robotic evaluation.}
 \label{fig:system_architecture_robot}
 \vspace{-0.35cm}
\end{figure}

The system Architecture for the \nameref{sec:robotic_evaluation} is shown in Fig. \ref{fig:system_architecture_robot}. This figure demonstrates how the robot substituted the user in Fig. \ref{fig:system_architecture_user} by the UR10 which takes the command to move from the impedance controller calculated from the laptop. The controller corrects the orientation and the linear speed of $z$ axis depending on the measured forces from the FT sensor.

\section{Data Acquisition Experiment}
\label{sec:Data_Acquisition_Experiment}

The goal of the data acquisition experiment is to record data from the participants (the movement of the charger and the FT sensor values) in order to obtain the needed parameters for the impedance controller: inertia $M_d$, damping $D_d$, and stiffness $K_d$.

\subsection{Experiment Design}

Before the experiment started, each user was explained the purpose of the experiment and demonstrated the operation. During the procedure, each user was asked to stand in front of the socket base and instructed to only carry the charger from the holder in order to obtain accurate readings from the FT sensor. Since phase two and phase three of the charging automation task do not depend on any CV operation but on the FT values, participants were asked to cover their eyes in order to maximize the dependence on their natural kinesthetic feedback on their hands.

\begin{figure}[h!]
 \centering
 \includegraphics[width=0.4\textwidth]{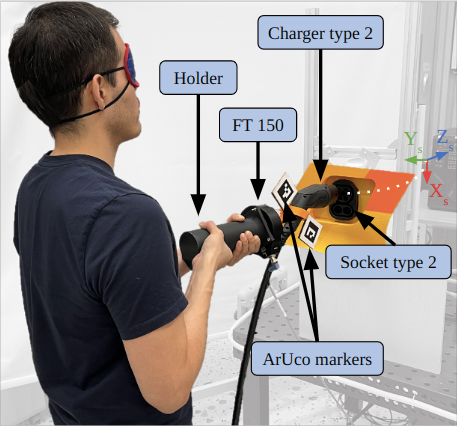}
 \caption{Data acquisition experimental setup.}
 \label{fig:experiment_setup}
 \vspace{-0.35cm}
\end{figure}

All participants started at a close position to the socket which allows them to start the plug-in phase without the need for searching for the socket. The initial orientation of the charger was not aligned with the orientation of the socket but was randomly tilted. The total error in the orientation was less than $10$ $deg$. The participants were supposed to depend on their kinesthetic feedback to correct the orientation of the charger while plugging in. The participants were also told when they had finished the plug-in phase so they could start with the plug-out phase. The procedure for this experiment is depicted in Fig. \ref{fig:experiment_setup}, in which one of the participants attempts to plug the charger into the socket.

\textbf{Participants:} Twenty-three participants, seven females and sixteen males, capable of performing the experiment, aged $24.5 \pm 2.4$ years, volunteered to participate in the experiment. All participants completed the experiment successfully.

\subsection{Data analysis}
\begin{figure}[]
 \centering
  \includegraphics[width=0.48\textwidth]{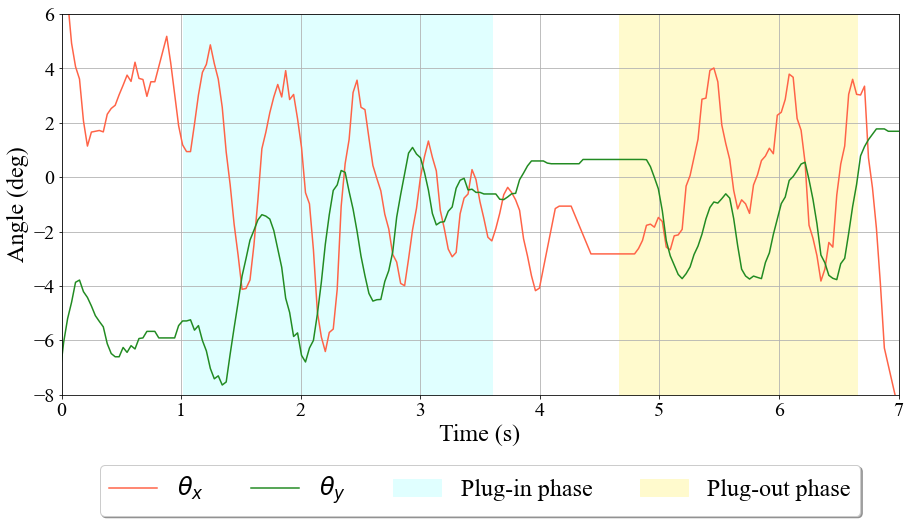}
 \caption{Changes of $\theta_x$ and $\theta_y$ during plug-in and plug-out phases for one of the users.}
 \label{fig:angle-changes-user}
 \vspace{-0.15cm}
\end{figure}

The movement of the charger was detected as the changes in the angles around $x_s$ and $y_s$ axes ($x$ and $y$ axes from the socket). These changes of the angles $\Delta{\theta_x}$ and $\Delta{\theta_y}$ were measured from the angle between $z_c$ axis (of the charger) and the $z_s$ axis (of the socket) projected on the $z_s-y_s$ plane and $z_s-x_s$ plane, respectively. Since the main concern of this work is to correct these two rotations, the changes in $F_x$ and $F_y$ were examined carefully and their changes were recorded and summarized in Table. \ref{table: max,mean,std all data}. The direction of the movement during the plug-in phase is in the positive direction of the $z_c$ axis and in the negative direction during the plug-out (can be concluded from Fig. \ref{fig:socket&charger}). This means that $F_z$ gives negative values while plugging in and positive values while plugging out (opposite sign to the direction of the movement). The response time of the charging process was determined according to the time needed for a user to change the direction of the charger. All data are summarized in Table \ref{table: max,mean,std all data} showing the maximum, mean, and standard deviation of the angles ($\Delta{\theta_x}, \Delta{\theta_y}$), forces ($\Delta F_x, \Delta F_y$), the force ($F_z$) and the response time during the whole task. We can see that the absolute value of the force to plug in the charger is higher than the force required to plug it out. This is due to the fact that the users remembered the path they should follow to accomplish the plug-out.
It was also noted that the $F_z$ for each user was remaining at the same value during the plug-in phase and at another constant value during the plug-out. An example of how the angles change can be seen in Fig. \ref{fig:angle-changes-user} where it shows $\theta_x$ and $\theta_y$ for one of the users where $\Delta{\theta_x} = 10.5$ $deg$, $\Delta{\theta_y} = 5.8$ $deg$, $t_{response} = 0.3 \: sec$.

\begin{table}[]
\centering
\caption{Maximum, mean, and standard deviation of the angles and forces applied by the participants using spiral strategy.}
\label{table: max,mean,std all data}
\renewcommand{\arraystretch}{1.3}
\begin{tabular}{lcccc}
\Xhline{4\arrayrulewidth}

 \multicolumn{1}{l}{$\mathbf{X}$}& \multicolumn{1}{c}{$\mathbf{X_{mean}}$} &\multicolumn{1}{c}{$\mathbf{X_{max}}$} & \multicolumn{1}{c}{$\mathbf{\boldsymbol{\sigma_{X}}}$} & \multicolumn{1}{c}{$\mathbf{Unit}$} \\ \Xhline{4\arrayrulewidth}
$\Delta\theta_x$ &  $9.5 $                               & $14.8$                              & $2.1 $   & $ deg$                                  \\ 
$\Delta\theta_y $ & $6.8 $                               & $11.3$                               & $1.8 $ & $deg$                                                               \\ 
$\Delta F_x $& $27.7 $                               & $49.8$                              & $10.3 $ & $N$                                                              \\ 
$\Delta F_y $ & $32.6$                               & $47.1$                               & $7.9 $ & $N$                                      \\ 
$F_{z\_ plug-in} $ & $-81.6 $                               & $-103.7 (min)$                            & $14.5 $ & $N$                                                                 \\ 
$F_{z\_ plug-out}$ & $ 75.6$                              & $90.1$                               & $8.6 $ & $N$                                   \\ 

$t_{response} $ &    $0.26$                            & $0.37$                               & $0.08 $ & $sec$                                                              \\ \Xhline{4\arrayrulewidth}
\end{tabular}
\end{table}

\section{Controller Design}
\label{sec:Controller_Design}
\begin{figure}[]
 \centering
  \includegraphics[width=0.48\textwidth]{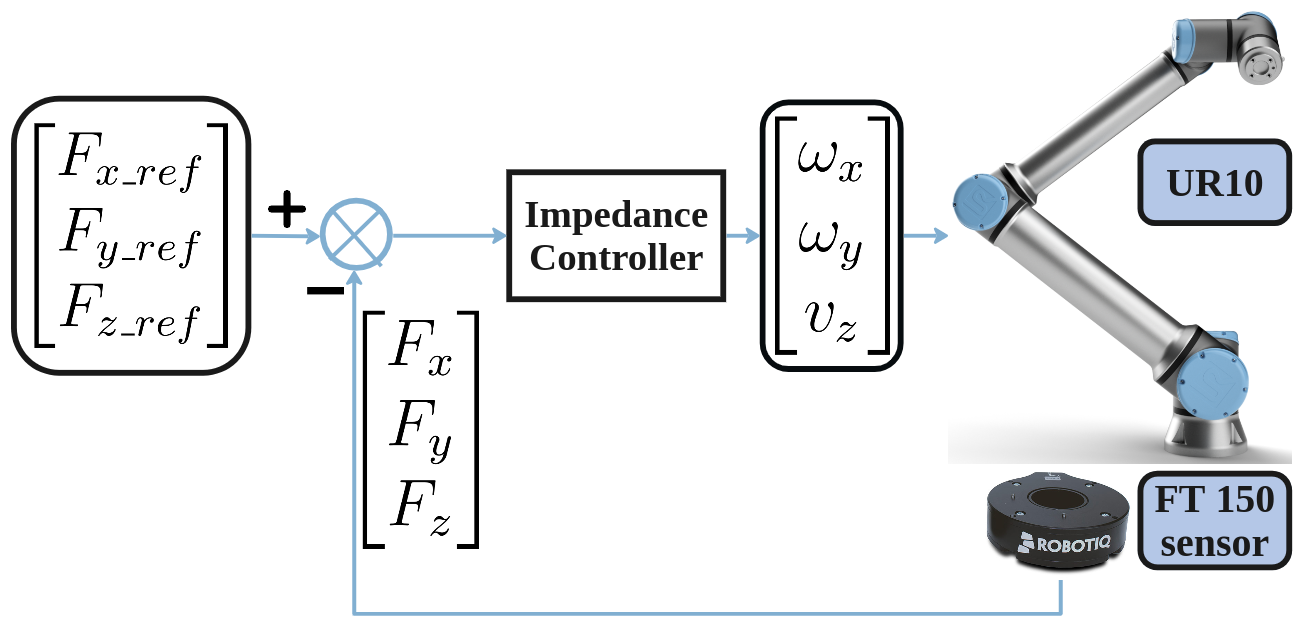}
 \caption{Diagram of the controller designed for the robot.}
 \label{fig:Controller}
\end{figure}

We propose to control the angular velocities of the charger (\nameref{sec:Rotational Movement}), $\dot{\theta}_x$ and $\dot{\theta}_y$ depending on the read forces, $F_x$ and $F_y$, respectively, and to control the linear velocity $v_z$ (\nameref{sec:Linear Movement}) depending on the reading force $F_z$ using impedance control. The diagram of the controller is shown in Fig. \ref{fig:Controller}, which consists of an impedance controller for the angular and linear velocity where the $F_{z\_ref}$ is the reference force depending on the phase of the charging process and $F_{x\_ref}$ and $F_{y\_ref}$ are zeros. The movement is relative to the coordinate system of the end effector of the charger.

\subsection{Rotational Movement}
\label{sec:Rotational Movement}
The angular velocities are controlled using impedance control, where the inputs of the controller are the forces $F_x$ and $F_y$. The angular motion of the charger around one axis can be expressed using the following equation:
\begin{equation}
\label{eq: motion around one axis}
    M_{d}\Ddot{\theta} + D_d\dot{\theta} + K_d\theta = F_{ext}(t),
\end{equation}
where $\theta$ is the angle of the end effector of the charger around one axis ($\theta_x$ or $\theta_y$), $M_d, D_d$, and $K_d$ are the desired parameters for moment of inertia, damping ration, and stiffness, respectively, and $F_{ext}$ is the force applied by the socket on the charger all on one of the axes.

The system in Eq. \ref{eq: motion around one axis} is a system of the second order that can be written in the Laplace domain as in Eq. \ref{eq: TF second order} \cite{transfer_function}:

\begin{equation}
\label{eq: TF second order}
H(s) = \frac{\Theta(s)}{F_{ext}(s)} = \frac{K_w \omega_n^2}{s^2 + 2\omega_n \zeta s + \omega_n^2},
\end{equation}
where $s$ is the Laplace operator variable, $K_w = \frac{1}{K_d} $ is the gain of the transfer function, $\omega_n^2 = \frac{K_d}{M_d}$ is the natural frequency and $\zeta = \frac{D_d}{2 M_d \omega_n^2 }$ is the damping ratio.

The desired response is with no overshoot ($\zeta = 1$), settle time $t_s = 0.2$ $sec$ and $K_w =  \frac{\Delta \theta}{\Delta F}$ is calculated depending on Table \ref{table: max,mean,std all data} and it corresponds to $K_w = 5.086 \times 10^{-3} \: rad/N$ and $4.285 \times 10^{-3} \: rad/N$ for the rotation around $x_c$ and $y_c$, respectively. Ultimately, we got the required parameters for our two controllers as seen in Table \ref{table: impedance parameters rotational}. The settling time was chosen to have a response faster than the human response that was recorded in Table. \ref{table: max,mean,std all data}.

\begin{table}[h]
\centering
\caption{Stiffness, Damping and Inertia Parameters of the impedance Controller for Rotational Movement around $x$ and $y$ axes.}
\label{table: impedance parameters rotational}
\renewcommand{\arraystretch}{1.3}
\begin{tabular}{cccc}
\Xhline{4\arrayrulewidth}

\multicolumn{1}{l}{$\mathbf{Parameter}$} & \multicolumn{1}{c}{$\mathbf x$ \textbf{axis}}        & \multicolumn{1}{c}{$\mathbf y$ \textbf{axis}}          &  \multicolumn{1}{c}{$\mathbf{Unit}$}\\ \Xhline{4\arrayrulewidth}

\multicolumn{1}{c}{$K_d$} & \multicolumn{1}{c}{$196.61$}        & \multicolumn{1}{c}{$233.30$}          &   $N.rad^{-1}$\\ 
\multicolumn{1}{c}{$D_d$} & \multicolumn{1}{c}{$19.66$}        & \multicolumn{1}{c}{$23.34$}          &  $N.sec.rad^{-1}$\\ 
\multicolumn{1}{c}{$M_d$} & \multicolumn{1}{c}{$0.4915$}        & \multicolumn{1}{c}{$0.5835$}          &  $N.sec^2.rad^{-1}$\\ \Xhline{4\arrayrulewidth}

\end{tabular}
\end{table}

\subsection{Linear Movement}
\label{sec:Linear Movement}
In the same way we derived the controller for the rotational movement, the impedance controller for the linear movement was designed. The linear motion of the charger on $z$ axis can be expressed using the following equation:
\begin{equation}
\label{eq: motion on z axis}
    M_{d}\Ddot{z} + D_d\dot{z} + K_d z = F_{z\_ext}(t),
\end{equation}
where $z$ is the position of the end effector on $z$ axis, $M_d, D_d$ and $K_d$ are the desired parameters for moment of inertia, damping ratio, and stiffness, respectively, and $F_{z\_ext}$ is the force applied by the socket on the charger on the same axis.

Reaching the final destination in the plug-in phase is determined by two factors, the measured force $F_z$ and the distance the robot had moved while inserting. In order to refrain from causing any damage to the socket, the controller was designed to follow the minimum absolute value between $F_{z\_plug-in}$ and $F_{z\_ plug-out}$ taken from Table \ref{table: max,mean,std all data}, we can call this value $F_{z\_ref}$. The controller regulates the force around $-75.6 \ N$ and $75.6 \ N$ during the plug-in and plug-out, respectively. The depth of the socket is $d_{depth} = 34.8 \ mm$ and at this value, the force reaches its minimum during the plug-in phase.

The desired response is with no overshoot ($\zeta = 1$), settle time $t_s = 0.2$ $sec$ and $K_w =  \frac{d_{depth}}{F_{z\_ref}} = 0.460\  mm/N$.
Ultimately, we got the required parameters for our controller as seen in Table \ref{table: impedance parameters linear}.

\begin{table}[h]
\centering
\caption{Stiffness, Damping and Inertia Parameters of the impedance Controller for Linear Movement on $z$ axis.}
\label{table: impedance parameters linear}
\renewcommand{\arraystretch}{1.3}
\begin{tabular}{ccc}
\Xhline{4\arrayrulewidth}

\multicolumn{1}{c}{\textbf{Parameter}} & \multicolumn{1}{c}{ \textbf{$\mathbf z$ axis}}          &   \textbf{Unit}\\ \Xhline{4\arrayrulewidth}

\multicolumn{1}{c}{$K_d$} & \multicolumn{1}{c}{$2.172$}          &   $N.mm^{-1}$\\ 
\multicolumn{1}{c}{$D_d$} & \multicolumn{1}{c}{$0.2172$}          &  $N.sec \ mm^{-1}$\\ 
\multicolumn{1}{c}{$M_d$} & \multicolumn{1}{c}{$5.431 \times 10^{-3}$}          &  $N.sec^2 \ mm^{-1}$\\ \Xhline{4\arrayrulewidth}

\end{tabular}
\end{table}

\section{Evaluation of Robot Performance}
\label{sec:robotic_evaluation}
\begin{figure}[]
 \centering
 \includegraphics[width=0.45\textwidth]{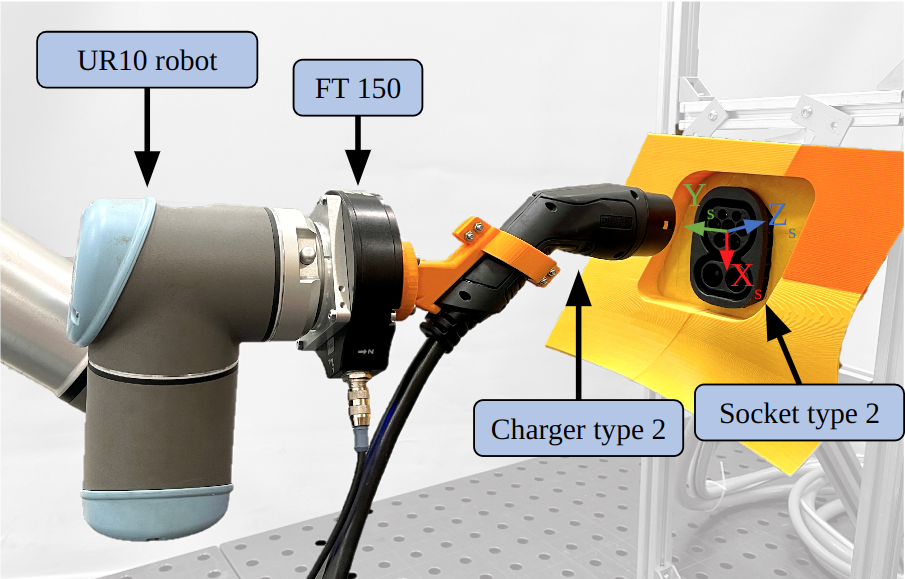}
 \caption{The charging setup used in the robotic evaluation.}
 \label{fig:charger_on_robot}
 \vspace{-0.15cm}
\end{figure}

To evaluate the performance of our controllers, the FT sensor and the charger were mounted on a UR10 robot and moved in front of the socket as it is shown in Fig. \ref{fig:charger_on_robot}. Similarly to the data acquisition experiment, the robot starts in a position near the socket where the plug-in phase can start without searching for the socket. The total error in the orientation between the charger and the socket was $10$ $deg$. We applied the designed controllers on the robot to plug in and plug out the electric charger. The results show that the designed controllers worked successfully and the movement of the robot and the values read by the FT sensor were recorded while plugging in and out.

\begin{figure}[]
 \centering
  \includegraphics[width=0.48\textwidth]{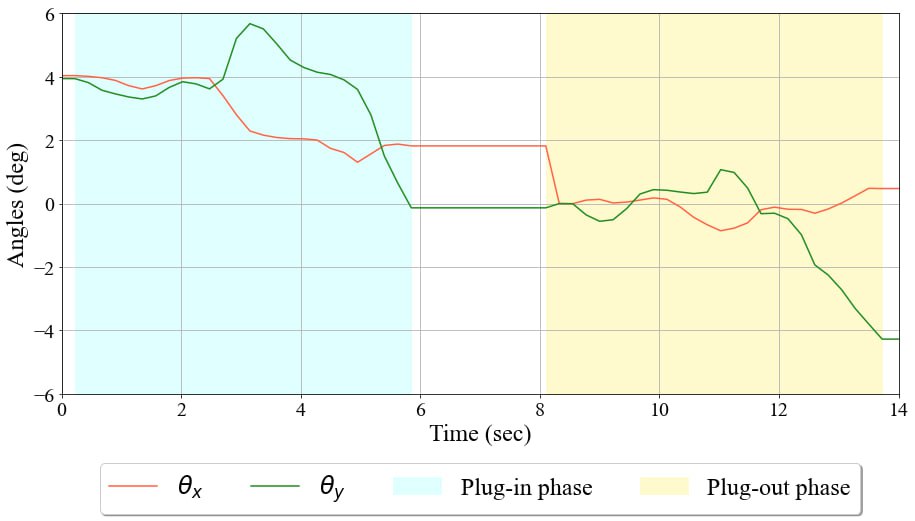}
 \caption{Changes of $\theta_x$ and $\theta_y$ during plug-in and plug-out phases for the robot.}
 \label{fig:angle-changes-robot}
 \vspace{-0.15cm}
\end{figure}

Fig. \ref{fig:angle-changes-robot} shows the changes in angles around $x$ and $y$ axes where the robot started from an initial orientation error of $4 \: deg$ for the two angles $\theta_x$ and $\theta_y$. The result of the plug-in phase shows an error smaller that $2 \: deg$ for both angles. During the plug-out phase, the angles keep changing until the phase is finished (determined by the $F_z$) then the robot stops.

\begin{figure}[]
 \centering
  \includegraphics[width=0.48\textwidth]{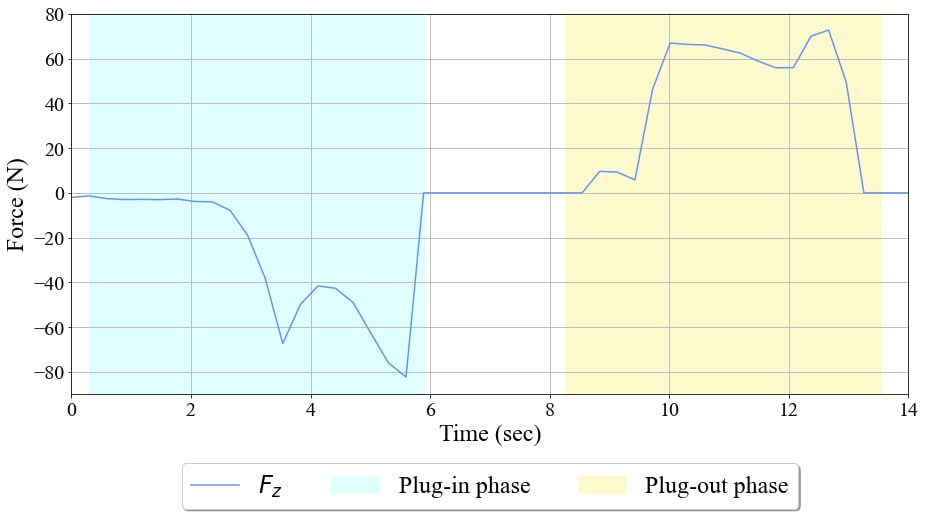}
 \caption{Changes of $F_Z$ during plug-in and plug-out phases for the robot.}
 \label{fig:force-changes-robot}
 \vspace{-0.35cm}
\end{figure}

The changes of the force $F_z$ through time can be observed in Fig. \ref{fig:force-changes-robot}. As seen from this figure the robot regulates the force around $F_{z\_ref}$ for each phase, around $-75.6 \ N$ and $75.6 \ N$ for the plug-in and plug-out, respectively.

\section{Conclusion and Future work}
\label{sec:conclusions}
In this work, we presented a novel method of plugging in and plugging out a robotic electric charger in an electric car socket. The final goal of this paper was to correct the rotations of the charger to align it with the socket and complete the plug-in and plug-out operations. Impedance Control was used to correct the angular velocities around two axes ($\dot{\theta}_x$ and $\dot{\theta}_y$) and the linear velocity ($v_z$). To calibrate the parameters of the controller, we depended on human haptic patterns. Participants were asked to perform the task of charging an electric car and their data were measured using an FT sensor and ArUco markers. A controller for plug-in and out based on impedance controller was designed and its parameters were determined based on human patterns. The proposed methodology and controller were evaluated in a robotic system, where the plug-in and out operation was performed successfully while the orientation error was smaller than $10 \ deg$.

Exceeding the limitation on orientation errors, which the system is capable to compensate, could prevent the system from operating properly. However, the orientation error of the socket in the first part of the automatic charging process using CV is less than $10 \ deg$, which makes our algorithm valid in almost all cases.

In future work, an extended study of the success of the proposed controller will be performed, where different position and orientation errors will be considered, with the aim to explore the limitations and improve the results. We aim to enhance the charging process by adding a correction to the rotation around the $z$ axis depending on $\tau_z$, and enhance the speed of the plug-in and out. An algorithm to cover the failure cases and how the robot should perform could be explored to enhance the performance of the controller.

The autonomous charger operation is considered to be one application of the peg-in-hole task. The algorithm applied in this research can be applied to other applications that include the peg-in-hole task, such as assembly tasks.

\addtolength{\textheight}{-12cm}   





\end{document}